\documentclass[11pt]{article}
\usepackage{coling2020}
\usepackage{times}
\usepackage{url}
\usepackage{latexsym}

\colingfinalcopy 

\usepackage{amsmath,amssymb,amsfonts}
\usepackage{algorithmic}
\usepackage{graphicx}
\usepackage{textcomp}
\usepackage{xcolor}

\usepackage{xspace}
\usepackage{multirow}
\usepackage{adjustbox}
\usepackage{framed}
\usepackage{tcolorbox}
\usepackage{bm}
\usepackage{subfig}
\usepackage{makecell}
\usepackage{blindtext}
\usepackage{booktabs}
\usepackage{url}
\usepackage{enumitem}
\usepackage{wrapfig}
\usepackage{tabularx}
\usepackage{floatrow}

\newtheorem{theorem}{Theorem}[section]
\newtheorem{definition}[theorem]{Definition}

\title{Dual Supervision Framework for Relation Extraction with Distant Supervision and Human Annotation}
\author{Woohwan Jung \\
  Seoul National University\\
  Seoul, Korea\\
  {\tt whjung@kdd.snu.ac.kr} \\\And
  Kyuseok Shim \\
  Seoul National University\\
  Seoul, Korea\\
  {\tt kshim@snu.ac.kr} \\}

\date{}

\mathchardef\mhyphen="2D 

\newcommand{\triple}[3]{\ensuremath{\langle #1,#2,#3\rangle}}
\newcommand{\minisection}[1]{\vspace{0.03in}{\bf \noindent #1.} }
\newcommand{\hanet}{\emph{HA-Net}\xspace}
\newcommand{\dsnet}{\emph{DS-Net}\xspace}
\newcommand{\munet}{$\mu$-Net\xspace}
\newcommand{\sigmanet}{$\sigma$-Net\xspace}

\newcommand{\lognormal}[2]{{L}(#1, #2)}
\newcommand{\head}{$e_{h}$\xspace}
\newcommand{\tail}{$e_{t}$\xspace}
\newcommand{\ehead}{e_{h}}
\newcommand{\etail}{e_{t}}
\newcommand{\norel}{\texttt{NA}\xspace}

\newcommand{\entity}[1]{\textbf{\color{blue!55!green}\textsf{#1}}}
\newcommand{\relation}[1]{\textbf{\color{red!55!yellow}\texttt{#1}}}
\newcommand{\relationt}[1]{\texttt{#1}}
\newcommand{\vect}[1]{\boldsymbol{\mathbf{#1}}}
\newcommand{\fracsmall}[2]{#1/#2}
\newcommand{\fracpas}[2]{(#1/#2)}
\newcommand{\transpose}[1]{{#1}^\top}
\newcommand{\realset}[1]{\mathbb{R}^{#1}}

\newcommand{\red}[1]{{\color{red} #1}}

\renewcommand{\figurename}{Figure}

\newcommand{\dcnn}{\emph{CNN\textsubscript{D}}\xspace}
\newcommand{\dlstm}{\emph{LSTM\textsubscript{D}}\xspace}
\newcommand{\dca}{\emph{CA\textsubscript{D}}\xspace}
\newcommand{\dbilstm}{\emph{BiLSTM\textsubscript{D}}\xspace}
\newcommand{\dbert}{\emph{BERT\textsubscript{D}}\xspace}

\newcommand{\sbilstm}{\emph{BiLSTM\textsubscript{S}}\xspace}
\newcommand{\scnn}{\emph{CNN\textsubscript{S}}\xspace}
\newcommand{\sbigru}{\emph{BiGRU\textsubscript{S}}\xspace}
\newcommand{\spcnn}{\emph{PCNN\textsubscript{S}}\xspace}
\newcommand{\spalstm}{\emph{PaLSTM\textsubscript{S}}\xspace}
\newcommand{\sbert}{\emph{BERT\textsubscript{S}}\xspace}

\newcommand{\dual}{\emph{DUAL}\xspace}

\newcommand{\bafix}{\emph{BAFix}\xspace}
\newcommand{\baset}{\emph{BASet}\xspace}
\newcommand{\maxth}{\emph{MaxThres}\xspace}
\newcommand{\entth}{\emph{EntThres}\xspace}
\newcommand{\dsonly}{\emph{DS-Only}\xspace}
\newcommand{\haonly}{\emph{HA-Only}\xspace}

\newcommand{\multitask}{\emph{Multitask}\xspace}
\newcommand{\single}{\emph{Single}\xspace}

\newcommand{\inmid}{\!\in\!}
\newcommand{\minusmid}{\!-\!}
\newcommand{\plusmid}{\!+\!}

\newcommand{\firstdef}[1]{\textbf{#1}}
\newfloatcommand{capbtabbox}{table}[][\FBwidth]

\newtcolorbox[auto counter]{somebox}[1][]{arc=0pt,auto outer arc,left=1pt,boxsep=1pt,boxrule=1pt,width=\columnwidth,right=1pt,#1}

\newcolumntype{C}{>{\Centering\arraybackslash}X} 

\graphicspath{{figures/}}
\DeclareGraphicsExtensions{.pdf,.jpeg,.png}

\begin{document}
\maketitle
\begin{abstract}
%

Relation extraction (RE) has been extensively studied due to its importance in real-world applications such as knowledge base construction and question answering.
Most of the existing works train the models on either distantly supervised data or human-annotated data.
To take advantage of the high accuracy of human annotation and the cheap cost of distant supervision, we propose the dual supervision framework which effectively utilizes both types of data.
However, simply combining the two types of data to train a RE model may decrease the prediction accuracy since distant supervision has labeling bias.
We employ two separate prediction networks \hanet and \dsnet to predict the labels by human annotation and distant supervision, respectively, to prevent the degradation of accuracy by the incorrect labeling of distant supervision.
Furthermore, we propose an additional loss term called \emph{disagreement penalty}
to enable \hanet to learn from distantly supervised labels.
In addition, we exploit additional networks to adaptively assess the labeling bias by considering contextual information.
Our performance study on sentence-level and document-level REs confirms the effectiveness of the dual supervision framework.


\end{abstract}
\section{Introduction}
\label{sec:intro}
Relation extraction (RE) has been widely used in real-world applications such as knowledge base construction \cite{dong2014knowledge,KnowledgeFusion,JungKS19}, question answering \cite{xu2016question} and biomedical data mining \cite{ahmed2019bio}.
Given a pair of entities in a text (e.g., sentence or document), the goal of RE is to discover the relationships between the entities expressed in the text.
More specifically, we aim to extract triples from the text in the form of \triple{\ehead}{r}{\etail} where \head is a head entity, \tail is a tail entity and $r$ is a relationship between the entities.

\blfootnote{This work is licensed under a Creative Commons Attribution 4.0 International License. License details: \url{http://creativecommons.org/licenses/by/4.0/}.}

To train a model for RE, we need a large volume of fully labeled training data in the form of text-triple pairs.
Although human annotation provides high-quality labels to train the relation extraction models, it is difficult to produce a large-scale training data since manual labeling is expensive and time-consuming.
Thus, \newcite{mintz2009distant} proposed distant supervision to automatically produce a large labeled data by using an external knowledge base (KB).
For a text with a head entity \head and a tail entity \tail, when a triple \triple{\ehead}{r}{\etail} exists in the KB for any relation type $r$, distant supervision produces a label \triple{\ehead}{r}{\etail} even though the relationship is not expressed in the text.
Thus, it suffers from the wrong labeling problem.
For instance, if a triple \triple{UK}{\relationt{capital}}{London} is in the KB, distant supervision labels the triple 
even for the sentence `London is the largest city of the UK'.


Although each of the two labeling methods has a certain weakness, most of the existing works for RE utilize either human-annotated (HA) data or distantly supervised (DS) data.
To take advantage of the high accuracy of human annotation and the cheap cost of distant supervision, we propose to effectively utilize a large DS data as well as a small amount of HA data.
Since DS data is likely to have \emph{labeling bias}, simply combining the two types of data to train a RE model may decrease the prediction accuracy.
To take a close look at the labeling bias, let the \emph{inflation} of a relation type be the ratio of the average frequencies of the relation type per text in DS data and HA data, respectively.
We say that a relation type is \emph{unbiased} if the average frequency of the relation type in DS data is the same as that in HA data (i.e., the inflation of the relation is 1).
By examining a document-level RE dataset (DocRED) \cite{yao2019docred} with 96 relation types, we found that the inflations of the relation types are from 0.48 to 85.9.
%
It indicates that distant supervision tends to generate a large number of false labels for some relation types.


Recently, \newcite{ye2019looking} introduced a domain adaptation approach to tackle the labeling bias problem for RE.
It trains a RE model on DS data and adjusts the bias term of the output layer by using HA data.
Although the bias adjustment achieves a meaningful accuracy improvement, it has a limitation.
An underlying assumption of the method is that the labeling bias is static for every text since it adjusts the bias term only once after training and uses the same bias during the test time.
However, the labeling bias varies depending on contextual information.
For example, in DocRED dataset, most of the \relationt{capital} relation labeled by distant supervision are false positive.
However, if the phrase `is the capital city of' appears in the text, the label is likely to be a true label.
Thus, we need to take account of contextual information to extract relations more accurately by considering the labeling bias.
To effectively utilize DS data and HA data for training RE models, we propose the \emph{dual supervision framework} that can be applied to most existing RE models to achieve additional accuracy gain.
Since the label distributions in HA data and DS data are quite different, we cast the task of training RE models with both data as a multi-task learning problem.
Thus, we employ the two separate output modules \hanet and \dsnet to predict the labels by human annotation and distant supervision, respectively, while previous works utilize a single output module.
This allows the different predictions of the labels for human annotation and distant supervision, and thus it prevents the degradation of accuracy by incorrect labels in DS data.
If we simply separate the prediction networks to apply the multi-task learning, \hanet cannot learn from distantly supervised labels.
To enable \hanet to learn from DS data, we propose an additional loss term called \emph{disagreement penalty}.
It models the ratio of the output probabilities from the prediction networks \hanet and \dsnet by using maximum likelihood estimation with log-normal distributions to generate the calibrated gradient to update \hanet to effectively reflect distantly supervised labels. 
Furthermore, our framework exploits two additional networks \munet and \sigmanet to adaptively estimate the log-normal distribution by considering contextual information.
Moreover, we theoretically show that the disagreement penalty enables \hanet to effectively utilize the labels generated by distant supervision.
Finally, we validate the effectiveness of the dual supervision framework on two types of tasks: sentence-level and document-level REs.
The experimental results confirm that our dual supervision framework significantly improves the prediction accuracy of existing RE models. 
In addition, the dual supervision framework substantially outperforms the state-of-the-art method \cite{ye2019looking} in both sentence-level and document-level REs with the relative F1 score improvement of up to 32\%.

\section{Preliminaries}
\label{sec:preliminaries}
We present the problems of sentence-level and document-level relation extractions and next introduce existing works for relation extraction.

\subsection{Problem Statement}
Following the works \cite{yao2019docred,wang2019fine}, we assume that each text is annotated with entity mentions.
For a pair of entities, since a sentence usually describes a single relationship between them, the sentence-level relation extraction is generally regarded as a \emph{multi-class} classification problem.
\begin{definition}[\textbf{Sentence-level relation extraction}] For a pair of the head and tail entities \head and \tail, a relation type set $R$ and a sentence $s$ annotated with entity mentions, we determine the relation $r \in R$ between \head and \tail in the sentence.
Note that $R$ includes a special relation type \norel which indicates that there does not exist any relation between \head and \tail.
\end{definition}

Since multiple relationships between a pair of entities can be expressed in a document,
document-level relation extraction is usually defined as a \emph{multi-label} classification problem.
\vspace{-0.02in}
\begin{definition}[\textbf{Document-level relation extraction}] For a pair of the head and tail entities \head and \tail, a relation type set $R$ and a document $d$ annotated with entity mentions, we find the set of all relations $R^* \subset R$ between \head and \tail appearing in document $d$. Note that $R$ does not include \norel in this case since it can be represented by an empty set of $R^*$.
\end{definition}

In this paper, we mainly discuss sentence-level RE and extend our framework to document-level RE.



\subsection{Existing Works of Relation Extraction}
\label{sec:existingworks}
\begin{figure}[tb]
	\subfloat[Existing models]{
		\includegraphics[width=2.3in]{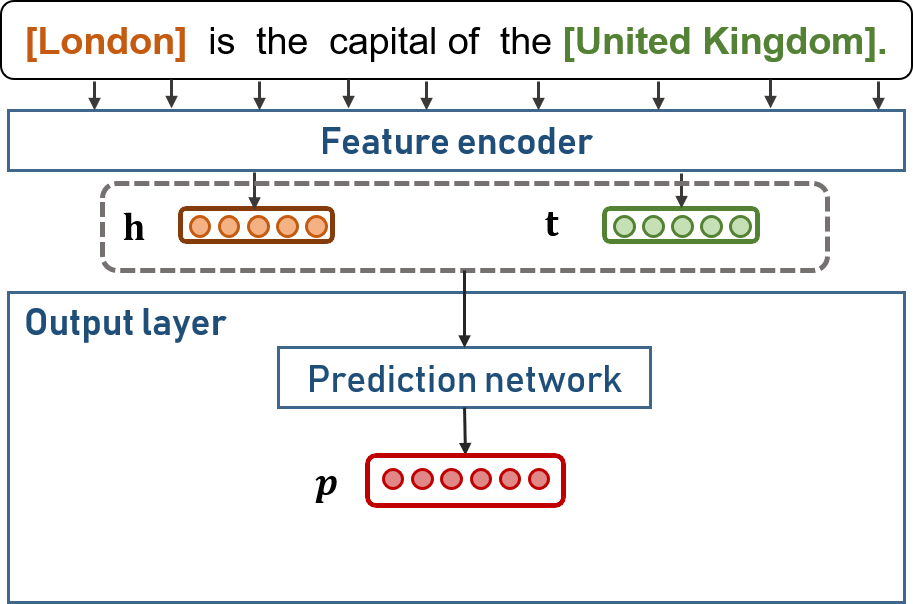}
		\label{fig:existing_framework}
	}
	\hspace{0.03in}
	\subfloat[The dual supervision framework]{
		\includegraphics[width=3.41in]{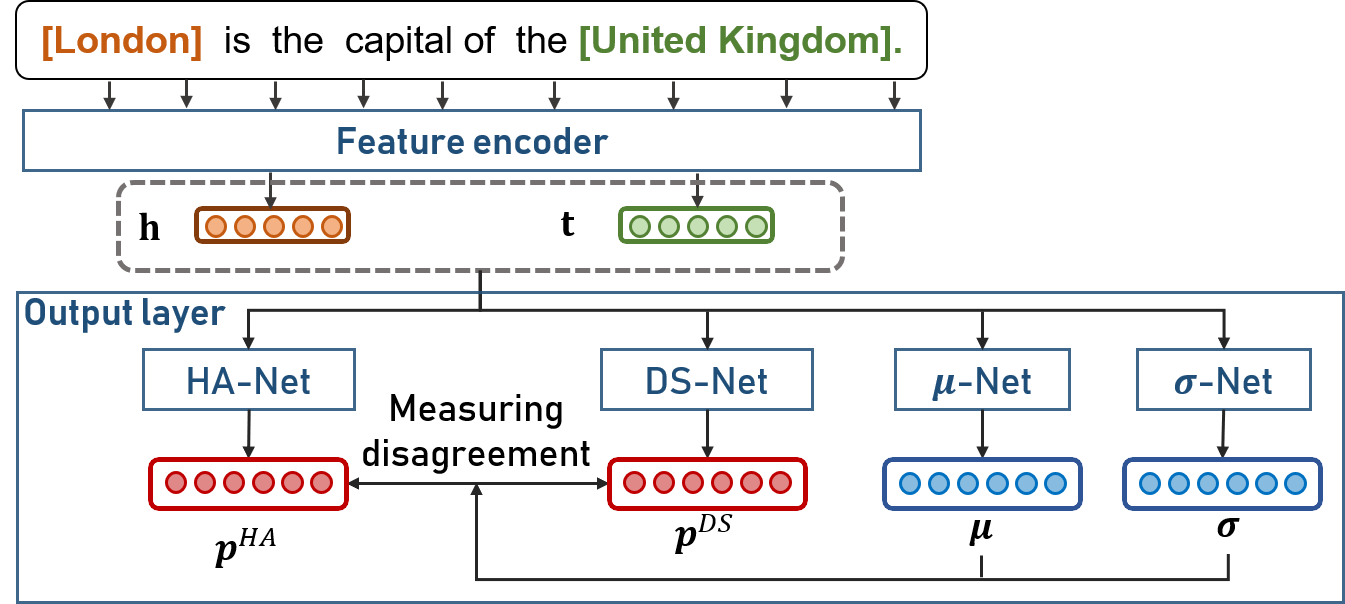}
		\label{fig:proposed_framework}
	}
	\vspace{-0.05in}
	\caption{The overall architectures of existing models and our framework}
	\label{fig:architecture}
	\vspace{-0.03in}
\end{figure}

A typical RE model consists of a feature encoder and a prediction network, as shown in \figurename~\ref{fig:architecture}\subref{fig:existing_framework}.
The feature encoder converts a text into the hidden representations of the head and tail entities. 
\newcite{cai2016bidirectional} and \newcite{wang2019fine} exploit Bi-LSTM and BERT, respectively, to encode the text.
On the other hand, \newcite{zeng2014relation} and \newcite{zeng2015distant} use CNN for the encoder. 
In addition, \newcite{zeng2014relation} propose the position embedding to consider the relative distance from each word to head and tail entities.  

The prediction network outputs the probability distribution of the relations between the entities.
Since sentence-level RE is a multi-class classification task, sentence-level RE models \cite{cai2016bidirectional,zeng2014relation,zeng2015distant} utilize a \emph{softmax classifier} as the prediction network and use categorical cross entropy as the loss function.
On the other hand, document-level RE models \cite{yao2019docred,wang2019fine} use a \emph{sigmoid classifier} and binary cross entropy as the prediction network and the loss function, respectively.
Since the labels obtained from distant supervision are noisy and biased,
with a single prediction network, it is hard to make accurate predictions for DS data and HA data together.

\section{Dual Supervision Framework}
We first present an overview of the dual supervision framework which effectively utilizes both human-annotated (HA) data and distantly supervised (DS) data for training RE models.
We next introduce the detailed structure of the output layer in our framework and propose our novel loss function with disagreement penalty that considers the labeling bias of distant supervision.
Then, we describe how to train the proposed model with both types of data as well as how to extract relations from the test data.
Finally, we discuss how the disagreement penalty makes each prediction network learn from the labels for the other prediction network although we use separate prediction networks.

\subsection{An Overview of the Dual Supervision Framework}
As shown in \figurename~\ref{fig:architecture}\subref{fig:proposed_framework}, our framework consists of a feature encoder and an output layer with 4 sub-networks.
It is general enough to accommodate a variety of existing RE models to improve their accuracy.
We can apply our framework to an existing RE model by using the feature encoder of the model and building the four sub-networks which exploit the structure of the original prediction network.
Since our framework uses the feature encoder of the existing models, we briefly describe only the output layer here.


Unlike the previous works, to allow the difference in the predictions for human annotated labels and distantly supervised labels, we exploit multi-task learning by employing two separate prediction networks \hanet and \dsnet to predict the labels in HA data and DS data, respectively.
We also use \hanet to extract relations from the test data.
The separation of the prediction networks prevents the accuracy degradation caused by incorrect labels from distant supervision.
If we simply utilize two prediction networks to apply the multi-task learning, \hanet cannot learn from distantly supervised labels although the prediction networks share the feature encoder.
To enable \hanet to learn from distantly supervised labels, we introduce an additional loss term called \emph{disagreement penalty}.
It models the disagreement between the outputs of \hanet and \dsnet by using maximum likelihood estimation with log-normal distributions.
Furthermore, to adaptively estimate the parameters of the log-normal distribution by considering contextual information, we exploit two parameter networks \munet and \sigmanet.

For a label \triple{\ehead}{r}{\etail}, 
let $I_{HA}$ be an indicator variable that is 1 if the label is obtained by human annotation and 0 otherwise.
The proposed framework uses the following loss function for a label \triple{\ehead}{r}{\etail}
\begin{equation}
\label{eq:loss_full}
L_{h, t} = I_{HA} \cdot L_{h,t}^{HA} + (1-I_{HA}) \cdot L_{h,t}^{DS} + \lambda \cdot L_{h,t}^{DS\mhyphen HA}
\end{equation}
where $L_{h,t}^{HA}$ and $L_{h,t}^{DS}$ denote the prediction loss of \hanet and \dsnet, respectively, and $L_{h,t}^{DS\mhyphen HA}$ is the disagreement penalty to capture the distance between the predictions by \hanet and \dsnet.
The hyper parameter $\lambda$ controls the relative importance of the disagreement penalty to the prediction errors.
By using a separate prediction network for each type of data and introducing the disagreement penalty, \hanet learns from distantly supervised labels while reducing overfitting to noisy DS data.

\subsection{Separate Prediction Networks}
To alleviate the accuracy degradation from the noisy labels in DS data, we utilize two prediction networks.
The network \hanet is used to predict the human-annotated labels from the train data and to predict relations from the test data.
The other prediction network  \dsnet predicts the labels obtained by distant supervision.
We use the prediction network of an existing model for both prediction networks of our framework without sharing the model parameters.
The prediction networks \hanet and \dsnet output the $|R|$-dimensional vectors $\vect{p}^{HA}=[p(r_1|\ehead,\etail,HA),...$ $,p(r_{|R|}|\ehead,\etail,HA)]$ and $\vect{p}^{DS} = [p(r_1|\ehead,\etail,DS),...$ $,p(r_{|R|}|\ehead,\etail,DS)]$, respectively, where $p(r|\ehead,\etail,HA)$ and $p(r|\ehead,\etail,DS)$ are the probabilities that there exists a label \triple{\ehead}{r}{\etail}, in HA data and DS data, respectively. 
We simply denote $p(r|\ehead,\etail,HA)$ and $p(r|\ehead,\etail,DS)$ by $p_r^{HA}$ and $p_r^{DS}$, respectively.

\subsection{Disagreement Penalty}
Distant supervision labels are biased and the size of the bias varies depending on the type of relation.
Moreover, the bias can vary depending on many other features such as the types of head and tail entities as well as the contents of a text.
Thus, we propose to use an effective disagreement penalty to model the labeling bias depending on the context where the head and tail entities are located.

\begin{wraptable}{R}{0.3\textwidth}
	\caption{The result of K-S test}
	\label{tab:distributions}
	\small
	\begin{tabular}{c|c}
		\toprule
		\textbf{Distribution}  & \textbf{p-value} \\
		\midrule
		Log-normal&{$0.008$}\\
		Weibull & $0.001$ \\
		Chi-square & $4.6 \times 10^{-10}$ \\
		Exponential & $3.6 \times 10^{-13}$ \\
		Normal& $1.2\times 10^{-15}$\\  
		\bottomrule
	\end{tabular}
\end{wraptable}
\noindent\textbf{Distribution of inflations.}
We measure the labeling bias by using the inflations of relations.
Recall that the inflation of a relation type is the ratio of the average frequencies of the relation type per text in DS data and HA data, respectively.
To investigate the distribution of inflations, we computed the inflations of 96 relation types in DocRED data.
Since Kolmogorov-Smirnov (K-S) test \cite{massey1951kolmogorov} is widely used to determine whether an observed data is drawn from a given probability distribution, we used it to find the best-fit distribution of the inflations.
Since the range of the inflation is $[0,\infty)$, we evaluated p-values of the four probability distributions supported on  $[0,\infty)$: Log-normal, Weibull, chi-square and exponential distributions. \
In addition, we include the normal distribution as a baseline.
Table~\ref{tab:distributions} shows the result of K-S test for DocRED data.
Note that a probability distribution has a high p-value if the probability distribution fits the data well.
Since the log-normal distribution has the highest p-value, it is the best-fit distribution among the five probability distributions.
Based on the observation, we model the disagreement penalty between the outputs of the two prediction networks. 

\minisection{Modeling the disagreement penalty}
We develop the disagreement penalty based on the maximum likelihood estimation.
Let $X_r$ be the random variable which denotes the ratio of $p^{DS}_{r}$ to $p^{HA}_{r}$. 
Since the inflation is the ratio of the number of labels in DS data and HA data,
the ratio $\fracsmall{p^{DS}_{r}}{p^{HA}_{r}}$ represents the \emph{conditional inflation} of the relation type $r$ conditioned on the text with head and tail entities.
Thus, we assume that $X_r$ follows a log-normal distribution $\lognormal{\mu_r}{\sigma_r^2}$ whose probability density function is
\begin{equation}
	\label{eq:lognormal}
	f(x) = \frac{1}{x \sigma_r \sqrt{2\pi}} exp\left(-\frac{(\log{x}-\mu_r)^2}{2 \sigma_r^2} \right).
\end{equation}

The disagreement penalty $L_{h,t}^{DS\mhyphen HA}$ is defined as the negative log likelihood of the conditional inflation ${p^{DS}_{r}}/{p^{HA}_{r}}$, which is obtained by substituting ${p^{DS}_{r}}/{p^{HA}_{r}}$ into Equation~\eqref{eq:lognormal} as follows:
\begin{equation}
\label{eq:disagreement}
\begin{aligned}
-\log{f\left({p^{DS}_{r}}/{p^{HA}_{r}}\right)} = \frac{1}{2}\left(\frac{\log{p^{DS}_{r}} - \log{p^{HA}_{r}}-\mu_r}{\sigma_r}\right)^2 +\log{p^{DS}_{r}} - \log{p^{HA}_{r}}+\log{\sigma_r} + \frac{\log{2\pi}}{2}.
\end{aligned}
\end{equation}
Since $\frac{\log{2\pi}}{2}$ is constant, we utilize the disagreement penalty in Equation~\eqref{eq:disagreement} without the constant term.

If we set $\mu_r$ and $\sigma_r$ to fixed values, we cannot effectively assess the conditional inflation since it can vary depending on the context.
For example, although the inflation of the relation type \relationt{capital} is high, the conditional inflation should be lower if a particular phrase such as `is the capital city of' appears in the text.
To take account of the contextual information,
we employ two additional networks \munet and \sigmanet to estimate the $\mu_r$ and $\sigma_r$ that are the parameters of log-normal distribution $\lognormal{\mu_r}{\sigma_r^2}$.

%

\subsection{Parameter Networks}
The parameter networks \munet and \sigmanet output the vectors $\vect{\mu}=[\mu_1,...,\mu_{|R|}]$ and $\vect{\sigma}=[\sigma_1,...,\sigma_{|R|}]$, respectively, which are
the parameters of the log-normal distributions to represent the conditional inflation for $r\in R$.
Both \munet and \sigmanet have the same structure as those of the prediction networks except their output activation functions.
For a log-normal distribution $\lognormal{\mu}{\sigma}$, the parameter $\mu$ can be positive or negative, and $\sigma$ is always positive.
Thus, we use a hyperbolic tangent function and a softplus function \cite{dugas2001incorporating} as the output activation functions of \munet and \sigmanet, respectively.


For example, if the prediction network of the original RE model consists of a bilinear layer and an output activation function,
the parameter vectors $\vect{\mu}\inmid \realset{|R|}$ and $\vect{\sigma}\inmid \realset{|R|}$ are computed from the head entity vector $\vect{h}\inmid \realset{d}$ and tail entity vector $\vect{t}\inmid \realset{d}$ as
\begin{equation*}
\vect{\mu} = tanh({\transpose{\vect{h}}\vect{W}^\mu\vect{t}+\vect{b}^\mu}), ~\vect{\sigma} = softplus({\transpose{\vect{h}}\vect{W}^\sigma\vect{t}+\vect{b}^\sigma}) + \varepsilon
\end{equation*}
where $softplus(x) = \log{(1+e^x)}$ and $\varepsilon$ is a sanity bound preventing extremely small values of $\sigma_r$ from dominating the loss function, and $\vect{W}^\mu \inmid \realset{d \times |R| \times d}$, $\vect{W}^\sigma \inmid \realset{d \times |R| \times d}$, $\vect{b}^\mu \inmid \realset{|R|}$ and $\vect{b}^\sigma \inmid \realset{|R|}$ are learnable parameters.
We set the sanity bound $\varepsilon$ to 0.0001 in our experiment.

\subsection{Loss Function}
For sentence-level relation extraction, we use the categorical cross entropy loss as the prediction losses $L_{h,t}^{HA}$ and $L_{h,t}^{DS}$.
For a label \triple{\ehead}{r}{\etail}, we obtain the following loss function from Equations \eqref{eq:loss_full} and \eqref{eq:disagreement}
\begin{equation}
\label{eq:loss_sent}
\vspace{-0.02in}
\begin{aligned}
L_{h, t} =&I_{HA} \cdot L_{h,t}^{HA} + (1-I_{HA}) \cdot L_{h,t}^{DS} + \lambda \cdot L_{h,t}^{DS\mhyphen HA} \\
=& -I_{HA} \cdot \log{p^{HA}_{r}} -(1-I_{HA})\log{p^{DS}_{r}} + \lambda \left[\frac{1}{2}\left(\frac{\ell_r-\mu_r}{\sigma_r}\right)^2+\ell_r+\log{\sigma_r}\right]\\
\end{aligned}
\end{equation}
where  $\ell_r=\log{p^{DS}_{r}} - \log{p^{HA}_{r}}$, and $I_{HA}$ is 1 if the label is from HA data and 0 otherwise.

\subsection{Analysis of the Disagreement Penalty}
\label{sec:analysis}
Let $\vect{w}_{HA}$ be a learnable parameter of \hanet which predicts relations in the test time.
We investigate the effect of the disagreement penalty by comparing the gradients of loss functions with respect to $\vect{w}_{HA}$ for a human annotated label and a distantly supervised label.

For a label \triple{\ehead}{r}{\etail}, let $\phi_r = \fracsmall{(\log{\fracpas{p^{DS}_{r}}{p^{HA}_{r}}} -\mu_r)}{\sigma_r^2}$.
If the label is human annotated, we obtain the following
gradient of the loss $L_{h,t}$ with respect to $\vect{w}_{HA}$ from Equation \eqref{eq:loss_sent}
\vspace{-0.03in}
\begin{equation}
\label{eq:gradha}
\vspace{-0.04in}
\begin{aligned}
\nabla L_{h,t} 
&= \nabla  L^{HA}_{h,t} + \vect{ 0} + \lambda \nabla  L^{DS\mhyphen HA}_{h,t}
= - \left(1+\lambda(1 \plusmid \phi_r)\right)\frac{1}{p_r^{HA}}\nabla p_r^{HA}.
\end{aligned}
\vspace{-0.03in}
\end{equation}

On the other hand, if the label is annotated by distant supervision,
the gradient becomes
\vspace{-0.03in}
\begin{equation}
\label{eq:gradds}
\vspace{-0.03in}
\begin{aligned}
\nabla L_{h,t} = \vect{ 0} + \vect{ 0} + \lambda \nabla  L^{DS\mhyphen HA}_{h,t}
= -\lambda\left(1+\phi_r\right)\frac{1}{p_r^{HA}}\nabla p_r^{HA}.
\end{aligned}
\vspace{-0.03in}
\end{equation}

The two gradients in Equations~\eqref{eq:gradha} and \eqref{eq:gradds} have the same direction of $-\nabla p_r^{HA}$.
It implies that a human annotated label and a distantly supervised label have similar effects on training \hanet
except that the magnitudes of gradients are calibrated by $1\plusmid\lambda(1 \plusmid \phi_r)$ and $\lambda(1 \plusmid \phi_r)$, respectively.
Thus, \hanet can learn from not only  human annotated labels but also distantly supervised labels by introducing the disagreement penalty.
Recall that the log-normal distribution $\lognormal{\mu_r}{\sigma_r}$ describes the conditional inflation for a given sentence with a head entity and a tail entity.
If the median $e^{\mu_r}$ of $\lognormal{\mu_r}{\sigma_r}$ has a high value, the distantly supervised label is likely to be a false label. 
Thus, we decrease the size of $\phi_r$ to reduce the effect of a distantly supervised label.
On the other hand, as the median $e^{\mu_r}$ becomes lower, the size of $\phi_r$ increases to aggressively utilize the distantly supervised label.

%

\subsection{Extension to Document-level Relation Extraction}
For the document-level RE, we use the \emph{binary} cross entropy as the prediction losses $L_{h,t}^{HA}$ and $L_{h,t}^{DS}$.
For a pair of entities \head and \tail, let $R_{h,t}$ be the set of relation types between the entities.
In the train time, we use the following loss function for document relation extraction
\vspace{-0.1in}
\begin{equation*}
\begin{aligned}
L_{h, t} =
&\minusmid I_{HA} \left(\sum_{r \in R_{h,t}}{\log{p^{HA}_{r}}} \plusmid \sum_{r \in R \setminus R_{h,t}}{\log{(1 \minusmid p^{HA}_{r})}} \right)\\
&\minusmid (1 \minusmid I_{HA})\left(\sum_{r \in R_{h,t}}{\log{p^{DS}_{r}}} \plusmid \sum_{r \in R \setminus R_{h,t}}{\log{(1\minusmid p^{DS}_{r})}} \right)
\plusmid \lambda \sum_{r \in R_{h,t}}{\left[\frac{1}{2}\left(\frac{\ell_r \minusmid \mu_r}{\sigma_r}\right)^2 \plusmid \ell_r \plusmid \log{\sigma_r}\right]}.
\end{aligned}
\vspace{-0.02in}
\end{equation*}
where  $\ell_r=\log{\fracsmall{p^{DS}_{r}}{p^{HA}_{r}}}$, and $I_{HA}$ is 1 if the labels are from HA data and 0 otherwise.
We obtain the same property shown in Section~\ref{sec:analysis} for the above loss function.
In the test time, we regard that the model outputs the triple $\triple{\ehead}{r}{\etail}$ if $p^{HA}_r$ is greater than a threshold which is tuned on the development dataset. 

\section{Experiments}
We conducted a performance study for sentence-level and document-level REs by following the experimental settings of \cite{ye2019looking} and \cite{yao2019docred,wang2019fine}, respectively.
All models are implemented in PyTorch and trained on a V100 GPU.
We initialized \hanet and \dsnet to have the same initial parameters.
More experimental details including implementations can be found in Appendix~\ref{sec:impl}.



 \begin{wraptable}{r}{0.54\textwidth}
 	\vspace{-0.2in}
 	\caption{Statistics of datasets}
 	\label{tab:dataset}
 	\footnotesize
 	\begin{tabular}{c|rrrr|r}
 		\toprule
 		\multirow{2}{*}{Data} & \multicolumn{4}{c|}{Number of instances} &\multirow{2}{*}{\makecell[c]{\# of rel.\\types}}\\
 		& Train-HA & Train-DS & Dev & Test \\
 		\midrule
 		\midrule
 		KBP  & 378 & 132,369 &14,103 &1,488& 7 \\  
 		NYT & 756 & 323,126 & 34,871 &3,021& 25 \\ 
 		\midrule
 		DocRED	& 38,269 & 1,508,320 & 12,332 &12,842& 96 \\ 
 		\bottomrule
 	\end{tabular}
 \end{wraptable}
\subsection{Experimental Settings}
\textbf{Dataset.}
KBP \cite{ling2012fine,ellis2012linguistic} and NYT \cite{riedel2010modeling,hoffmann2011knowledge} are datasets for sentence-level RE, and DocRED \cite{yao2019docred} is a dataset for document-level RE.
The statistics of the datasets are summarized in Table~\ref{tab:dataset}.
Since KBP and NYT do not have HA train data, 
we use 20\% of the HA test data as the HA train data.
In addition, we randomly split 10\% of train data on KBP and NYT for the development (dev) data.
Note that the ground truth of the test data in DocRED is not publicly available.
However, we can get the F1 score of the result extracted from the test data by submitting the result to the DocRED competition hosted by CodaLab (available at https://competitions.codalab.org/competitions/20717).
We report both the F1 scores computed from the dev data and the test data.

\minisection{Compared methods}
We compare our dual supervision framework, denoted by \firstdef{\dual}, with the state-of-the-art methods \textbf{\baset} and \textbf{\bafix} in \cite{ye2019looking}.
For sentence-level RE, we compare \dual with two additional baselines \textbf{\maxth} \cite{ren2017cotype} and \textbf{\entth} \cite{liu2017heterogeneous} which are only applicable to multi-class classification and cannot be used in document-level RE.
\maxth outputs \norel if the maximum output probability is less than a threshold. Similarly, \entth outputs \norel if the entropy of the output probability distribution is greater than a threshold. 

\minisection{Used relation extraction models} 
For \emph{sentence-level RE}, we used the six models: \textbf{\sbigru}  \cite{zhang2017position}, \textbf{\spalstm} \cite{zhang2017position}, \textbf{\sbilstm} \cite{zhang2017position}, \textbf{\scnn} \cite{zeng2014relation}, \textbf{\spcnn} \cite{zeng2015distant} and \textbf{\sbert} \cite{wang2019fine}.
On the other hand, for \emph{document-level RE}, we used the five models:
\textbf{\dbert} \cite{wang2019fine}, \textbf{\dcnn} \cite{zeng2014relation}, \textbf{\dlstm} \cite{yao2019docred}, \textbf{\dbilstm} \cite{cai2016bidirectional} and \textbf{\dca} \cite{sorokin2017context}.
Note that \dcnn, \dbilstm, and \dca are originally proposed for sentence-level RE 
and we used the adaptation of them to document-level RE by \newcite{yao2019docred}.
In addition, we adapt \dbert to the sentence-level RE by changing the output activation function from sigmoid to softmax and denote it by \sbert.

\begin{table}[tb]
	\center
	\caption{Sentence-level RE datasets (KBP and NYT)\label{tab:performance_sentence}}
	\scriptsize
	\begin{tabular}{c|rrrrrr|rrrrrr}
\toprule
Dataset  &  \multicolumn{6}{c|}{KBP}  &  \multicolumn{6}{c}{NYT}\\
RE models & \sbigru & \spalstm & \sbilstm & \spcnn & \scnn & \sbert & \sbigru & \spalstm & \sbilstm & \spcnn & \scnn & \sbert\\
\midrule
\midrule
\haonly & 0.1984 & 0.1153 & 0.1787 & 0.3410 & 0.2586 & 0.1631 & 0.0884 & 0.1259 & 0.1504 & 0.4463 & 0.3978 & 0.1953\\
\dsonly & 0.3909 & 0.3521 & 0.3519 & 0.2705 & 0.2810 & 0.3610 & 0.4532 & 0.4429 & 0.4297 & 0.4177 & 0.4463 & 0.4625\\
\midrule
\baset & 0.3972 & 0.4055 & 0.4053 & 0.2410 & 0.2400 & 0.3858 & 0.4966 & 0.4555 & 0.4561 & 0.3584 & 0.4358 & 0.5081\\
\bafix & 0.4241 & 0.4027 & 0.3581 & 0.2931 & 0.2473 & 0.3383 & 0.4613 & 0.4507 & \textbf{0.4707} & 0.4023 & 0.4532 & 0.5145\\
\maxth & 0.4264 & 0.3630 & 0.4053 & 0.2815 & 0.2645 & 0.3751 & 0.4531 & 0.4462 & 0.4350 & 0.4258 & 0.4655 & 0.4952\\
\entth & 0.4470 & 0.4018 & \textbf{0.4248} & 0.2925 & 0.2826 & 0.3539 & 0.4553 & 0.4472 & 0.4210 & 0.4154 & 0.4427 & 0.4940\\
\midrule
\dual & \textbf{0.4749} & \textbf{0.4420} & 0.4207 & \textbf{0.3872}  & \textbf{0.2969} & \textbf{0.4013}& \textbf{0.5455} & \textbf{0.5210} & 0.4524 & \textbf{0.4986} & \textbf{0.4744}& \textbf{0.5300}\\
\bottomrule
	\end{tabular}
\end{table}

\subsection{Comparison with Existing Methods}
We compare the dual supervision framework with the existing methods. 

\minisection{Sentence-level RE}
\tablename~\ref{tab:performance_sentence} shows F1 scores for relation extraction on KBP and NYT.
Note that \dsonly and \haonly represent the original RE models trained only on distantly supervised and human-annotated labels, respectively.
\dual shows the highest F1 scores with all RE models except \sbilstm.
Since KBP and NYT have a small number of human-annotated labels in train data, \haonly shows worse F1 scores than \dsonly.
Furthermore, \dual achieves improvements of F1 score from 5\% to 40\% over \dsonly by additionally using the small amount of human annotated labels.
On the other hand, the compared methods \bafix, \baset, \maxth and \entth often perform worse than \dsonly and \haonly.


\begin{table}[tb]
	\center
	\caption{Document-level RE dataset (DocRED) \label{tab:performance_doc}}
	\scriptsize
	\begin{tabular}{c|ccccc|ccccc}
\toprule
   &   \multicolumn{5}{c|}{Dev}   &   \multicolumn{5}{c}{Test}  \\
\midrule
RE models  &   \dbert  &   \dbilstm  &   \dca  &   \dlstm  &   \dcnn  &   \dbert  &   \dbilstm  &   \dca  &   \dlstm  &   \dcnn\\
\midrule
\midrule
\haonly&0.5513&0.4992&0.4986&0.4817&0.4788&0.5478&0.4982&0.4992&0.4815&0.4681\\
\dsonly&0.4683&0.4951&0.4890&0.4877&0.4166&0.4587&0.4809&0.4772&0.4713&0.4160\\
\midrule
\baset&0.4807&0.5123&0.5024&0.5012&0.4349&0.4716&0.4949&0.4905&0.4905&0.4320\\
\bafix&0.4802&0.5136&0.5070&0.5166&0.4365&0.4730&0.5061&0.4989&0.4977&0.4354\\
\midrule
\dual&\textbf{0.5880}&\textbf{0.5510}&\textbf{0.5372}&\textbf{0.5392}&\textbf{0.4967}&\textbf{0.5774}&\textbf{0.5379}&\textbf{0.5306}&\textbf{0.5277}&\textbf{0.4909}\\
\bottomrule
	\end{tabular}
\end{table}
\minisection{Document-level RE}
We present F1 scores on DocRED in \tablename~\ref{tab:performance_doc}.
\dual outperforms \baset and \bafix with all RE models.
Especially, the F1 score of dual framework with \dbert shows more than 22\% of improvement over \baset and \bafix.
Since DocRED has a large human-annotated train data, \haonly shows better performance than \dsonly.
For \dbert and \dcnn, the existing methods show lower F1 scores compared to \haonly.
It shows that the accuracy can be degraded although we use additional DA data in addition to HA data due to the labeling bias.
Meanwhile, we achieve a consistent and significant improvement by applying \dual.
In the rest of this paper, we will provide a detailed evaluation of performance on DocRED data which is the largest dataset in this experiment.
For the test data of DocRED, the ground truth is not publicly available and only a F1 score can be obtained from the DocRED competition.
Thus, we provide detailed evaluations of performance on the dev data only.

\begin{figure*}[tb]
	\hspace{-0.14in}
	\subfloat[\dbert]{\includegraphics[width=3.2in]{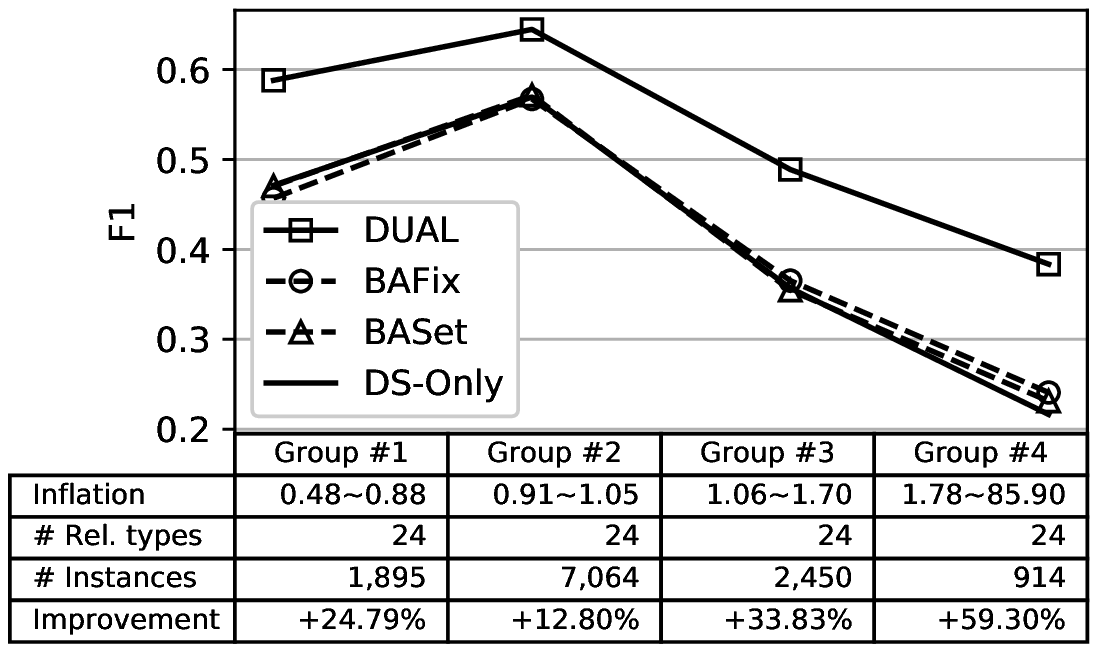}\label{fig:acc_by_inf_bert}}
	\hspace{-0.16in}
	\subfloat[\dbilstm]{\includegraphics[width=3.2in]{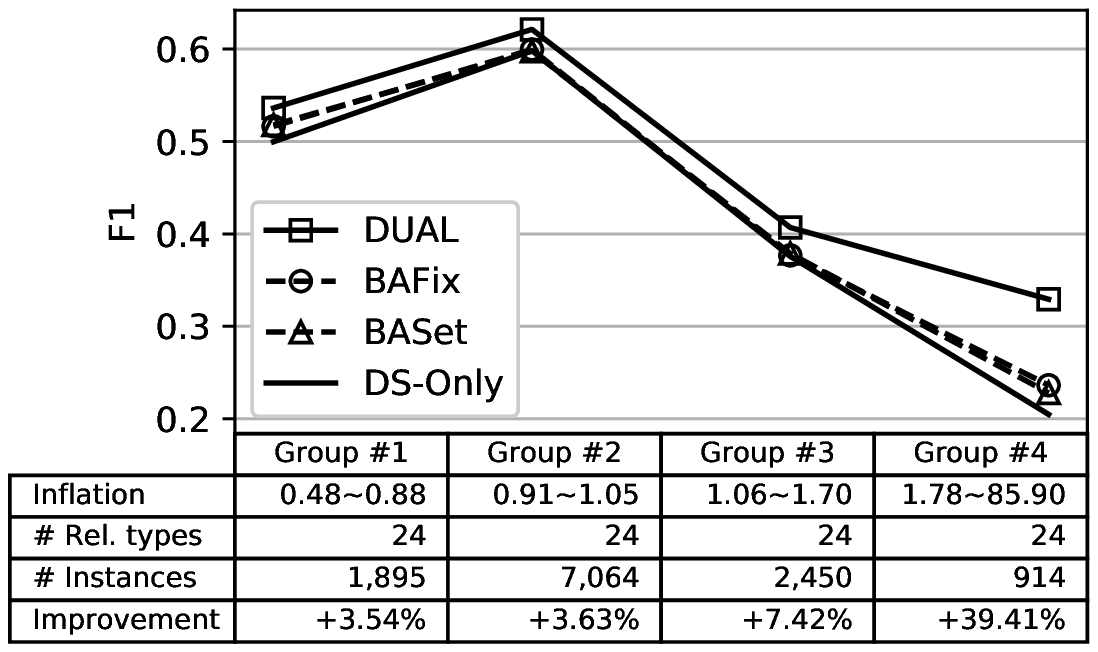}\label{fig:acc_by_inf_bilstm}}
	\vspace{-0.03in}
	\caption{F1 scores of different groups \label{fig:acc_varying_inf}}
	\vspace{-0.08in}
\end{figure*}

\minisection{Inflation vs. accuracy}
To investigate the effect of the inflation to the accuracy of relation extraction,
we split the relation types into 4 groups based on the inflation of the relation types.
In \figurename~\ref{fig:acc_varying_inf}, we present the characteristics of each group and plot the F1 scores by groups for \dbert model and \dbilstm model.
All methods have the highest F1 scores when the inflation is close to 1 (at the 2nd group).
Furthermore, the improvement of F1 score by \dual compared to the second best performer increases as the inflation moves away from 1.
Thus, it confirms that our dual supervision framework effectively utilizes both human annotation and distant supervision by modeling the bias of the distant supervision.
Since the other models \dca, \dlstm and \dcnn show similar results with \dbilstm, we omit the result.



\begin{wrapfigure}{R}{0.4\textwidth}
	\vspace{-0.16in}
	\includegraphics[width=1\textwidth]{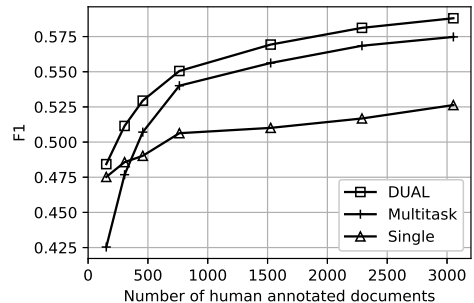}
	\vspace{-0.13in}
	\caption{Varying the size of HA data \label{fig:acc_by_had}}
	\vspace{-0.13in}
\end{wrapfigure}
\subsection{Ablation Study}
We conducted an ablation study with the existing model \dbert on DocRED to validate the effectiveness of individual components of our framework.
We compared \dual (separate prediction networks + disagreement penalty) and two variations of our framework \multitask (separate prediction networks only) and \single.
\multitask denotes a variation of \dual which does not utilize the disagreement penalty,
while  \dbert without applying the dual supervision framework  is referred to as \single.
Note that \single is also trained on both HA data and DS data together.


To show the effectiveness of the components depending on the size of HA data, we plotted the F1 scores with varying the number of human-annotated document from 152 to 3,053 (i.e., from 5\% to 100\% of the documents with HA).
As we expected, \dual outperforms both variations all the time.
Furthermore, separation of the prediction networks significantly improves the accuracy when we have enough number of human-annotated labels.
However, when we use less than 10\% of the human annotated documents, \multitask suffers from the sparsity problem.
By utilizing the disagreement penalty additionally, \dual outperforms \single even when we use only 5\% of the human-annotated documents for training the model.
It implies that the disagreement penalty enables \hanet to effectively learn from DS data as well as HA data.

\subsection{Quality Comparison}
To give an idea of what false relations are found by existing methods,
we provide two example documents in the dev data of DocRED and the relations extracted by \dual, \bafix and \dsonly with \dbert in \tablename~\ref{tab:casestudy}.
The relation \triple{Sweden}{\relationt{capital}}{Stockholm} is expressed in the document titled `Kungliga Hovkapellet' and
all methods find the relation correctly.
In the document titled `Loopline Bride', the relation \triple{Ireland}{\relationt{capital}}{Dublin} does not exist.
However, \bafix and \dsonly output the incorrect relation.
Since \dual adaptively assess the labeling bias with \munet and \sigmanet, \dual does not output the false relation.
In addition, since the RE models trained with \bafix and \dsonly fail to learn the text pattern corresponding to the relation type due to the labeling bias, they output many false labels such as \triple{Vietnam}{\relationt{capital}}{Taipei} in many documents.
It shows that the dual supervision framework effectively deal with the labeling bias of distant supervision by considering contextual information.

\begin{table}[tb]
	\setlength\extrarowheight{2pt}
	\small
	\caption{Examples of documents and extracted relations \label{tab:casestudy}}
	\begin{tabularx}{\textwidth}{c|X|X}
		\toprule
		\multirow{1}{*}{Document}  & Title: Kungliga Hovkapellet  & Title: Loopline Bridge\\
		& 
		[1] Kungliga Hovkapellet is a Swedish orchestra, originally part of the Royal Court in \entity{[Sweden]}'s capital \entity{[Stockholm]}.
		[2] Its existence ...
		&
		[1] The Loopline Bridge (or the Liffey Viaduct) is a railway bridge spanning the River Liffey and several streets in \entity{[Dublin]}, \entity{[Ireland]}.
		[2] It joins ...\\
		\midrule
		\multirow{1}{*}{Relations} 
		& True label: \triple{\entity{Sweden}}{\relation{capital}}{\entity{Stockholm}}&  True label:\norel \\
		& \dual: \triple{\entity{Sweden}}{\relation{capital}}{\entity{Stockholm}}&  \dual:\norel \\
		& \bafix: \triple{\entity{Sweden}}{\relation{capital}}{\entity{Stockholm}}& \bafix: \triple{\entity{Ireland}}{\relation{capital}}{\entity{Dublin}} \\
		& \dsonly: \triple{\entity{Sweden}}{\relation{capital}}{\entity{Stockholm}}&  \dsonly:\triple{\entity{Ireland}}{\relation{capital}}{\entity{Dublin}} \\
		\bottomrule
	\end{tabularx}
	\vspace{-0.04in}
\end{table}

\begin{wraptable}{r}{0.29\textwidth}
	\vspace{-0.2in}
	\caption{Topic-aware RE}
	\label{tab:topic_dual}
	\small
	\begin{tabular}{c|cc}
		\toprule
		& F1 & AUC	\\	
		\midrule
		\haonly & 0.6569 & 0.6456\\
		\dsonly & 0.6624 & 0.6978\\
		\dual & \textbf{0.6930}& \textbf{0.7125}\\
		\bottomrule
	\end{tabular}
\end{wraptable}

\subsection{Topic-aware RE}
Topic-aware RE is a special case of document-level RE to extract the relations between the topic entity of a document and the other entities.
\newcite{jung2020trex} proposed a topic-aware relation
extraction (T-REX) model which is robust to the omitted mentions
of topic entities in documents.
We apply our dual supervision framework to the T-REX on DocRED dataset and report the result in \tablename~\ref{tab:topic_dual}.
The result shows that our dual supervision framework is also effective in the topic-aware RE task.

\section{Related Works}
We briefly survey the existing works for RE.
\newcite{mintz2009distant} propose distant supervision to overcome the limitation of the quantity of human-annotated labels.
They utilize lexical, syntactic and named entity tag features obtained by existing NLP tools to extract relations.
Other early works in \cite{riedel2010modeling,hoffmann2011knowledge} also utilized hand-crafted features to find the relations in text.
However, since such RE models take the input features from NLP tools, the errors generated by the NLP tools are propagated to the RE models.
In order to deal with the error propagation, as we discussed in Section~\ref{sec:existingworks}, the works \cite{lin2016neural,zeng2014relation,zeng2015distant,sorokin2017context,wang2019fine} use deep neural networks such as CNN, LSTM and BERT instead of handcrafted features to encode the text for finding the relations.
Since many relational facts are expressed across multiple sentences, the recent works \cite{yao2019docred,wang2019fine} studied document-level RE.
\newcite{yao2019docred} provide a document-level RE dataset (DocRED)  as well as compare the models adapted from the sentence-level RE models \cite{zeng2014relation,hochreiter1997long,cai2016bidirectional,sorokin2017context}.
Moreover, a fine tuned model \cite{wang2019fine} of BERT \cite{devlin2018bert} for document-level RE achieved a higher F1 score than the baselines on DocRED.


The wrong labeling problem in distant supervision has been addressed in many previous works \cite{zeng2015distant,lin2016neural,ye2019distant,beltagy2018combining}.
Among them, \newcite{zeng2015distant}, \newcite{lin2016neural} and \newcite{ye2019distant} 
build a bag-of-sentences for a pair of entities and extract relational facts from the bag-of-sentences with attention over the sentences.
\newcite{beltagy2018combining} propose a bag-of-sentences-level model which utilizes human annotation.
However, they use the human annotated labels only to determine whether there exists a relationship or not since the labels are obtained from a different domain.
The goal of these works is different from ours which is to find the relations \emph{appearing in a given text} (e.g., a document). 
Thus, the bag-of-sentences-level models have a limitation to be used for some applications such as question answering.



The most relevant work to ours is \cite{ye2019looking}.
This paper proposes the bias adjustment methods to utilize a small amount of HA data to improve RE models trained on DS data 
by considering the different distribution of human annotated labels and distantly supervised labels.
However, they do not use HA data to train the models and use the HA data only to obtain a statistic to be used the determine the size of the bias adjustment.
Thus, the bias adjustment methods cannot consider contextual information.

\section{Conclusion}
We proposed the dual supervision framework to utilize human annotation and distant supervision based on the analysis of labeling bias in distant supervision.
We devised a new structure for the output layer of RE models that consists of 4 sub networks.
The new structure is robust to the noisy labeling of distant supervision since the labels obtained by human annotation and distant supervision are predicted by separate prediction networks \hanet and \dsnet, respectively.
In addition, we introduced an additional loss term called \emph{disagreement penalty} which enables \hanet  to learn from distantly supervised labels.
The parameter networks \munet and \sigmanet adaptively assess the labeling bias by considering contextual information.
Moreover, we theoretically analyzed the effect of the disagreement penalty.
Our experiments showed that the dual supervision framework significantly outperforms the existing methods.

\section*{Acknowledgements}
This work was supported by Next-Generation Information Computing Development Program through the National Research Foundation of Korea(NRF) funded by the Ministry of Science, ICT (No. NRF-2017M3C4A7063570)
and was also supported by Institute of Information \& communications Technology Planning \& Evaluation(IITP) grant funded by the Korea government(MSIT) (No. 2020-0-00857, Development of cloud robot intelligence augmentation, sharing and framework technology to integrate and enhance the intelligence of multiple robots).
This research was results of a study on the "HPC Support" Project, supported by the ‘Ministry of Science and ICT’ and NIPA.
\bibliographystyle{coling}
\bibliography{ref}
\newpage
\appendix
\section*{Appendix}

\section{Experimental Details}
\label{sec:impl}
Our implementation is available at \url{https://github.com/woohwanjung/dual}. 

\minisection{Document-level RE}
For \dbilstm, \dlstm, \dca and \dcnn, we utilized the code which is available at \url{https://github.com/thunlp/DocRED} and implemented by \newcite{yao2019docred}.
In addition, we used the implementation of \dbert that is available at \url{https://github.com/hongwang600/DocRed} and provided by  \newcite{wang2019fine}.
We used Adam optimizer \cite{kingma2014adam} to optimize the RE models.
For the \dbert model, we set the batch size to 12 and learning rate to $10^{-5}$.
For the other models, we followed the setting provided in \cite{yao2019docred}: batch size is 40, learning rate is $10^{-3}$. 
We set the hyperparameters $\lambda$ and $d$ to $10^{-5}$ and 128, respectively. 
Each training batch has half of the instances with human-annotated labels and the other half of instances with distantly supervised labels.

\minisection{Sentence-level RE}
We use the code which is made publicly available by \newcite{ye2019looking} at \url{https://github.com/INK-USC/shifted-label-distribution}.
All models except \sbert are trained by stochastic gradient descent.
Learning rate is initially set to 1.0, and decreased to 10\% if there is no improvement on the dev data for 3 consecutive epochs. 
For the models, we set the hyperparameters $\lambda$ and $d$ to $10^{-3}$ and 200, respectively.
To train \sbert model, we used Adam optimizer with learning rate $10^{-5}$.
Moreover, the hyperparameters $\lambda$ and $d$ are set to $10^{-4}$ and 128, respectively.
We alternately used an HA batch and a DS batch for dual supervision where an HA batch consists of training instances with human annotated labels and a DS batch consists of training instances with distantly supervised labels.


\section{Additional Experiments}

\begin{figure}[tb]
	\subfloat[\dbert]{\includegraphics[width=2.8in]{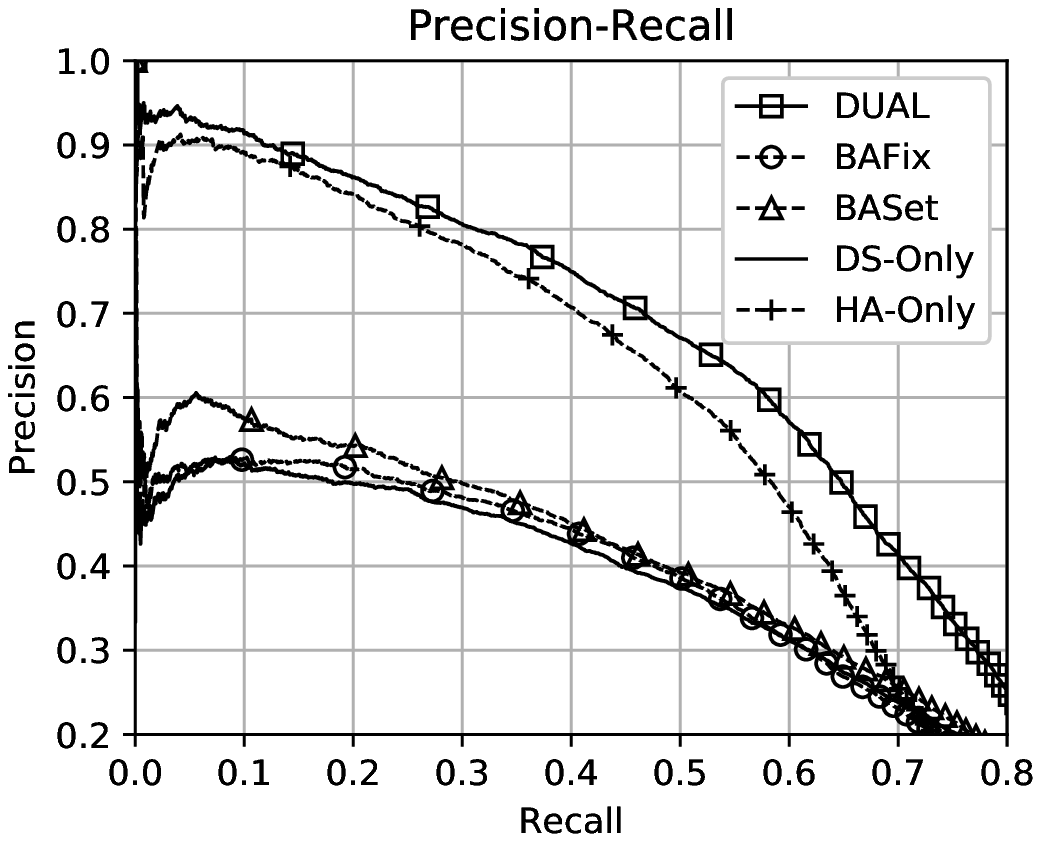}\label{fig:pr_bert}}
	\subfloat[\dbilstm]{\includegraphics[width=2.8in]{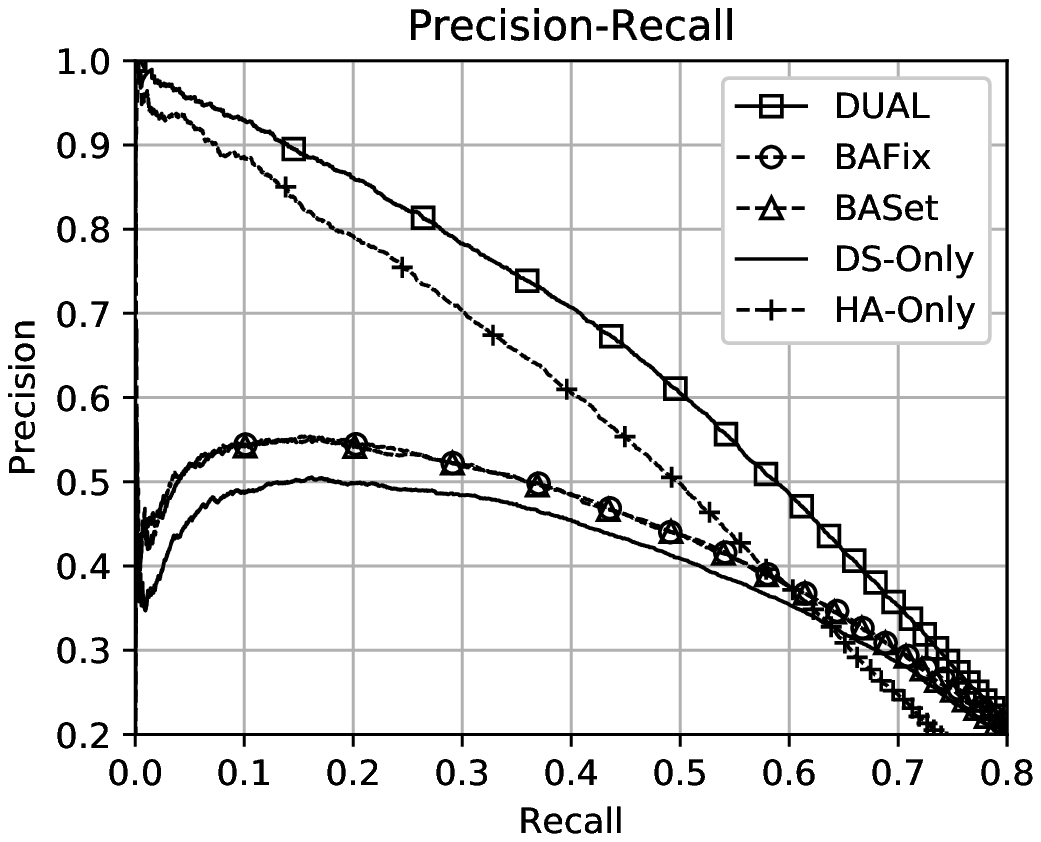}\label{fig:pr_bilstm}}
	\caption{Precision-recall curves \label{fig:pr_curves}}
\end{figure}

The precision-recall curves of the compared methods are shown in \figurename~\ref{fig:pr_curves}.
As expected, \dual consistently outperforms all compared methods.
\bafix and \baset have similar precision-recall curves with \dsonly.
Although \haonly shows comparable precisions with \dual when recall is low,
the precision of \haonly drops faster than that of \dual with increasing recall.
It implies that human annotated labels are not enough for training a model to extract a large number of relations.
Meanwhile, \dual extracts more relations from the document compared to existing models at the same precision level.

\end{document}